\documentclass{article}

\usepackage[final]{nips_2017}

\usepackage[utf8]{inputenc} 
\usepackage[T1]{fontenc}    
\usepackage{hyperref}       
\usepackage{url}            
\usepackage{booktabs}       
\usepackage{amsfonts}       
\usepackage{nicefrac}       
\usepackage{microtype}      
\usepackage{graphicx}

\title{Neurology-as-a-Service for the Developing World}

\author{
  Tejas Dharamsi, Payel Das, Tejaswini Pedapati, Gregory Bramble, \\\bf{Vinod Muthusamy, Horst Samulowitz, and Kush R. Varshney} \\
  IBM Research AI\\
  Thomas J. Watson Research Center\\
  Yorktown Heights, NY 10598 \\
  \AND
  Yuvaraj Rajamanickam, John Thomas, and Justin Dauwels \\
  School of Electrical and Electronic Engineering \\
  Nanyang Technological University \\
  Singapore 639798
}

\begin{document}

\maketitle

\begin{abstract}
  Electroencephalography (EEG) is an extensively-used and well-studied technique in the field of medical diagnostics and treatment for brain disorders, including epilepsy, migraines, and tumors. The analysis and interpretation of EEGs require physicians to have specialized training, which is not common even among most doctors in the developed world, let alone the developing world where physician shortages plague society. This problem can be addressed by teleEEG that uses remote EEG analysis by experts  or by local computer processing of EEGs. However, both of these options are prohibitively expensive and the second option requires abundant computing resources and infrastructure, which is another concern in developing countries where there are resource constraints on capital and computing infrastructure. In this work, we present a cloud-based deep neural network approach to provide decision support for non-specialist physicians in EEG analysis and interpretation.  Named `neurology-as-a-service,' the approach requires almost no manual intervention in feature engineering and in the selection of an optimal architecture and hyperparameters of the neural network. In this study, we deploy a pipeline that includes moving EEG data to the cloud and getting optimal models for various classification tasks.  Our initial prototype has been tested only in developed world environments to-date, but our intention is to test it in developing world environments in future work.  We demonstrate the performance of our proposed approach using the BCI2000 EEG MMI dataset, on which our service attains 63.4\% accuracy for the task of classifying real vs.\ imaginary activity performed by the subject, which is significantly higher than what is obtained with a shallow approach such as support vector machines.
\end{abstract}

\section{Introduction}

Classification of EEGs is an important application used in various fields, such as brain function monitoring, medical diagnostics for epilepsy, seizures, and brain computer interfaces (BCI). EEG detects voltage fluctuations at the scalp, which is related to the neuronal activity in the brain, over time. A sophisticated setup that includes advanced computing infrastructure and trained physicians is often needed on-site for EEG interpretation for diagnostics and treatment. EEG is most often used to diagnose  epilepsy that causes spikes or abnormal activities in the EEG recordings. According to WHO reports [1], epilepsy is a disorder more prominent in developing countries than developed countries and majority of patients in developing countries do not receive necessary treatment due to lack of skilled physicians and computing resources, resulting in a so-called treatment gap. Lack of training and skill among physicians can also lead to misdiagnosis of epilepsy, resulting in adverse outcomes. 

One way to automate the EEG-based classification is to use machine learning techniques. In the past, several machine learning techniques have been proposed for EEG-based feature learning/classification of brain states by carefully engineering features from the signals. An experienced and skillful scientist with domain knowledge is often required for hand engineering  the features,  as these features from the EEG recordings  may not generalize to different subjects or tasks/situations. Moreover, it is a time consuming process.

Deep Neural Networks are characterized for EEG feature learning [8],[9]. Also, with the advent of cloud computing in providing cost-effective way of performing software-as-a-service, we propose an economical way for analysis of EEGs in the cloud using deep neural networks to help physicians make better diagnoses.  The patient's EEG recording trials are moved to the cloud after data acquisition. The collected data goes through a minimal preprocessing step involving bandpass filtering and power spectogram generation which uses short term Fourier transform changing the signal into frequency time domain. This 3D data with electrodes, frequency and time are fed to convolutional neural network with optimal hyperparameter setting. The results of the classification task are returned to the physician for his analysis to better his judgment. In this study, we describe our results on task classification  on a BCI EEG dataset. Our motivation is to create and test a system that can be utilized by physicians around the world with minimal human intervention.

\begin{figure}
    \centering
    \includegraphics[width=55ex,height=45ex]{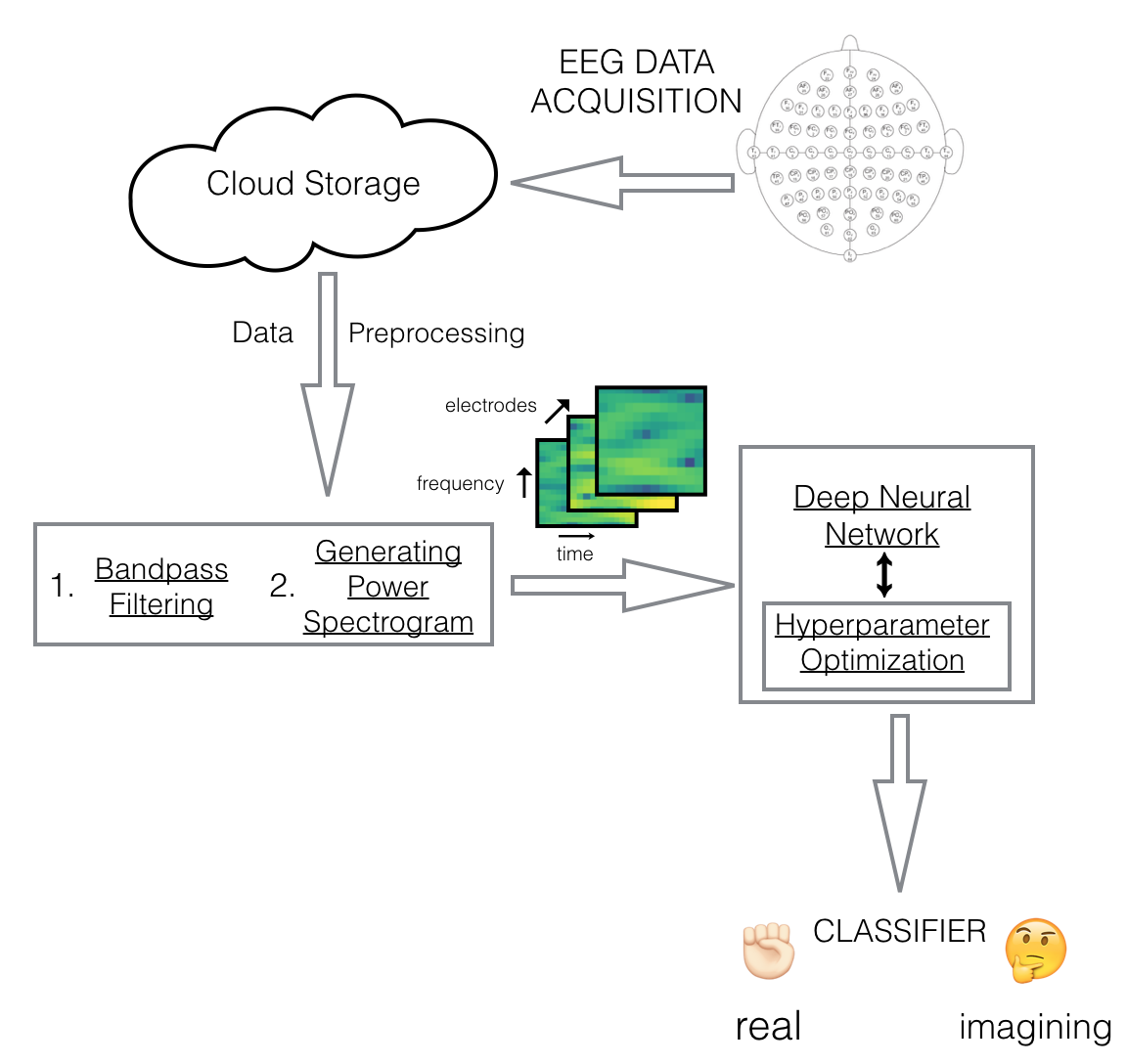}
    \caption{Flow Diagram}
\end{figure}
\section{Method}

In our experiments, we utilized Physionet's [2] BCI2000 EEG MMI dataset [3] that consists of 1500 trials from 109 subjects, who performed tasks such as opening or closing of their fists or feet, or doing both, or imagining doing so. Data was collected using 64 electrodes as per a 10-10 international system as shown in Figure 1 above. The sampling frequency for EEG data acquisition was 160 Hz. Out of 14 tasks to be performed by each subject, we considered 12 tasks which involved the subject to either perform a movement or imagine a movement.  We used EEGs for 103 subjects in the study since 6 subjects had data annotations different from the dataset description.

Data preprocessing of the raw EEG signals is an essential task prior to feeding it to any machine learning system due to presence of noise and artifacts.  Therefore, we applied a bandpass filter with the  range of 3 Hz - 30 Hz on the raw EEG data to filter out uninteresting frequency ranges. In order to treat each annotated trial segment (4.1 seconds long) as a sequence of stacked images, we generate spectrogram associated with each channel in the input trial with time segments of length 0.8 sec and sliding window of 0.05 sec. We used hanning window and NFFT of size 128 for spectogram computation.  We averaged the power values over a frequency band of 2.5 Hz for the entire frequency range and  computed the relative log power for each time segment. Relative power (RP) for i\textsuperscript{th} frequency band is defined as real power (P) in a frequency band over sum of real log power in all the frequency band.  
\begin{center}
    {$RP_i$ = $P_i$ / $\sum_{i=1}^{n}$ $P_i$}.
\end{center}
These aforementioned pre-processing steps are standard signal processing tasks and can be applied as an automated process to all incoming EEG recordings without any manual intervention. This pre-processing resulted in a 3D representation of the EEG input data, in which the dimensions are electrode channels, frequency bands, and time segments (Figure 1). 

The prepared data on cloud was fed to a convolutional neural network (CNN). We utilized an automated mechanism to select optimal hyperparameters for CNN's architecture [4].   In our study, we used random optimizer [6],[7] with the mechanism mentioned above. Parameters, such as number of layers, learning rate, type of layers, number of filters, and their size for the convolutional layer, probability range for dropouts were explored. Input to the optimizer were lower bound and upper bound for each of those hyperparameters. This model selection mechanism uses  IBM "Deep Learning as a Service" [5], a cloud-based service to run deep learning models. The pipeline returned a model with best validation accuracy for the classification task at hand.

\section{Results}

In our study, to create end-to-end pipeline for `neurology-as-a-service,'  we performed a classification task to determine if the movement of a subject is real or imaginary using the dataset discussed above. The prepared data consisted of 17232 samples and was split at the ratio of 7:3 for training and testing. A random optimizer was used for hyperparameter fine-tuning. The service returned a best accuracy of 63.4\% by choosing the optimal model after 750 iterations. The layer configuration of the optimal model had 3 hidden layers: convolutional (conv) layer with 61 filters of size 5x5, a second conv layer with 69 filters of size 8x8, and a max-pooling layer (pool) with filter size 5x5 applied with a stride of 2 and learning rate of 0.001. Figure 2 shows the accuracy obtained using different hyper-parameter configurations as a function of iteration step using the random optimizer. Zero accuracy in some iterations indicates that the hyperparameter chosen by the optimizer did not result in a feasible network.

We also employed the hand-crafted 3D CNN model used in EEG classification reported in ref [10], which comprises of 3 pairs of conv (filter size 3x3) and max-pooling layer (2x2, stride 2) with number of filters in conv layer being 32, 64, and 128 respectively followed by a fully-connected (fc) layer with size 512. This model returned an accuracy of 58.21\% on validation set. The training accuracy obtained was 95.39\%,  suggesting that the automatically optimized 3DCNN model performs better in terms of overfitting compared to a hand-crafted architecture. In addition, our approach generates several different models yielding comparable performance. 

In order to compare our results, we also ran a standard SVM on a 500-dimensional space, as obtained by  performing  principal component analysis (PCA) on the pre-processed data, which resulted in an accuracy of 56\%. Thus, higher prediction accuracy obtained using a CNN model, in which hyperparameters are automatically fine-tuned using an optimizer, shows the potential of our proposed framework. 

Table 1 below shows the top five model configurations. Layer configurations are indicated with layer type and their corresponding attributes like number of filters, filter size, keep-probability are shown within parentheses.

\begin{table}
\begin{center}
 \begin{tabular}{||c|c|c||} 
 \hline
 No. of hidden layers & Layer Configuration & Best Accuracy  \\ [0.5ex] 
 \hline\hline
 3 &  conv(61,5x5), conv(69,8x8), pool(5x5,2) & 63.4 \\ 
 \hline
 4 &  conv(210,5x5), fc(828), dropout(0.71), fc(18) & 62.23 \\ 
 \hline
 1 & fc(2266) & 62.2  \\
 \hline
 2 & fc(664), fc(1025) & 62.0   \\
 \hline
 2 & pool(4x4, 2), conv(247,11x11) & 61.9  \\
 \hline
 
\end{tabular}

\caption{Top five 3DCNN model configurations and their accuracy}

\end{center}
\end{table}
 
\begin{figure}
    \centering
    \includegraphics[width=55ex,height=35ex]{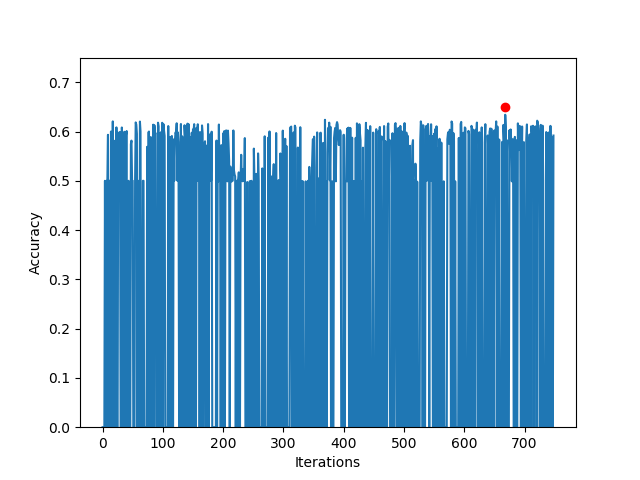}
    \caption{Variation in accuracy over optimizer iterations}
\end{figure}

\section{Conclusion and Future Work}
For the task of EEG classification we have proposed a cost- and time-efficient framework that requires minimal human intervention, which uses deep neural networks processed over cloud computing infrastructure and can be deployed in developing countries for diagnostics and treatment of brain disorders. 

As part of our next steps, we plan to use this framework on a dataset aimed at classification of epileptic seizures and/or pathological/normal EEG. We further aim to test the framework in developing work environment and improve the existing framework accordingly. We would also like to see how the framework performs using other hyperparameter optimization techniques including Bayesian optimization. 

\subsubsection*{Acknowledgments}

This work was conducted under the auspices of the IBM Science for Social Good initiative.

\section*{References}

\medskip

\small

[1] \url{http://www.who.int/mediacentre/factsheets/fs999/en/}

[2] Goldberger AL et al., "PhysioBank, PhysioToolkit, and PhysioNet: Components of a New Research Resource for Complex Physiologic Signals." Circulation 101(23):e215-e220 [Circulation Electronic Pages; http://circ.ahajournals.org/cgi/content/full/101/23/e215]; 2000 (June 13).

[3] Schalk, G. et al., "BCI2000: A General-Purpose Brain-Computer Interface (BCI) System." IEEE Transactions on Biomedical Engineering 51(6):1034-1043, 2004.

[4] Diaz G. I., Fokoue-Nkoutche A., Nannicini G. and Samulowitz H., "An effective algorithm for hyperparameter optimization of neural networks", IBM Journal of Research and Development 61(4):9, 2017

[5] Bhattacharjee B. et al., "IBM Deep Learning Service", IBM Journal of Research and Development 61(4):10, 2017

[6] Bergstra J. et al., "Algorithms for Hyper-Parameter Optimization", Advances in Neural Information Processing Systems 24: 2546--2554, 2011

[7] Bergstra J. and  Bengio Y. "Random search for hyper-parameter optimization", Journal of Machine Learning Research 13:281-305, 2012

[8] Bashivan P., Rish I., Yeasin M. and Codella N., "Learning Representations from EEG with Deep Recurrent-Convolutional Neural Networks", arXiv preprint arXiv:1511.06448, 2015

[9] Schirrmeister, R. T., Springenberg, J. T., Fiederer, L. D. J., Glasstetter, M., Eggensperger, K., Tangermann, M., Hutter, F., Burgard, W. and Ball, T., "Deep learning with convolutional neural networks for EEG decoding and visualization", Human  Brain Mapping 38(11): 5391–5420, 2017

[10] Hung Y., Wang Y., Prasad M., Lin C.,"Brain Dynamic States Analysis based on 3D Convolutional Neural Network", IEEE International Conference on Systems, Man, and Cybernetics (SMC), 2017.

\end{document}